\pdfoutput=1
\documentclass[12pt,a4paper]{article}

\usepackage[utf8]{inputenc}
\usepackage[T1]{fontenc}
\usepackage{lmodern}
\usepackage[english]{babel}

\usepackage{setspace}
\usepackage{parskip}

\usepackage{amsmath}
\usepackage{amsfonts}
\usepackage{amssymb}
\usepackage{mathtools}

\usepackage{graphicx}
\usepackage{float}
\usepackage{subcaption}

\usepackage{tikz}
\usetikzlibrary{positioning,shapes,arrows,arrows.meta,calc,fit,backgrounds}

\usepackage{pgfplots}
\pgfplotsset{compat=1.18}

\usepackage{booktabs}
\usepackage{array}
\usepackage{multirow}
\usepackage{tabularx}

\usepackage{algorithm}
\usepackage{algorithmic}

\usepackage[backend=biber,style=authoryear,url=true,doi=false,eprint=false,maxnames=99,maxcitenames=2]{biblatex}

\usepackage[colorlinks=true, linkcolor=blue, citecolor=blue, urlcolor=blue]{hyperref}
\usepackage{cleveref}

\usepackage{listings}
\usepackage{geometry}

\AtEveryBibitem{%
  \clearfield{note}%
  \clearfield{pagetotal}%
  \clearfield{pages}%
  \clearfield{isbn}%
  \clearfield{issn}%
  \clearfield{archivePrefix}%
  \clearfield{eprint}%
  \clearfield{primaryClass}%
  \clearname{editor}%
  \clearfield{urldate}%
}

\DeclareFieldFormat{citehyperref}{%
  \DeclareFieldAlias{bibhyperref}{noformat}%
  \bibhyperref{#1}}

\DeclareCiteCommand{\cite}[\mkbibparens]
  {\usebibmacro{prenote}}
  {\usebibmacro{citeindex}%
   \printtext[citehyperref]{\usebibmacro{cite}}}
  {\multicitedelim}
  {\usebibmacro{postnote}}

\renewcommand{\mkbibparens}[1]{\mkbibbrackets{#1}}

\title{Artificial Intelligence as a Tool for Combating Child Labour:\\
A Real-Time Edge Vision Pipeline for Child Detection and Age Estimation}
\author{Mark Nowak\\Conflux Laboratory\\\texttt{mnowak@confluxlab.org}}
\date{\today}

\begin{document}

\maketitle
\thispagestyle{empty}

\begin{abstract}
An estimated 138 million children remain in child labour worldwide, and the monitoring systems the affected sectors rely on, built on periodic household visits and interviews, systematically under-detect them.
We present a real-time computer-vision pipeline, built and operated solely as a research prototype, that studies the feasibility of giving such systems, in particular Child Labour Monitoring and Remediation Systems (CLMRS), a continuous presence-based evidence channel.
The pipeline combines a multi-task person and face detector (a YOLO26x backbone within the CerberusDet framework), cascaded age and gender estimation that pairs MiVOLO~v2 with a child-specialist model for ages 0--12, ByteTrack tracking, ArcFace and DINOv2 re-identification, and track-level fusion that outputs per-person records with confidence tiers and quality flags for human review.
The detector raises person mAP@0.5 from 0.390 to 0.683 over the previous-generation baseline on a 13,537-image benchmark.
The child specialist reaches a mean absolute error of 1.944 years on children-only validation (ages 0--12); on identical child-face crops, widely used open-source stacks err by 18.7 to 22.9 years, systematically ageing children into adults.
We document the training protocol including its failures, and the mathematical formulation of the deployed models, fusion rule and identity-consolidation constraints.
FP8 TensorRT compilation accelerates the age cascade 1.77-fold at a cost of 0.002 years MAE, bringing the pipeline above twice real-time speed on embedded Blackwell-class hardware.
On 26.8 hours of proxy video the system surfaces 634 unique child candidates against 285 for its predecessor, including 9.3 times more children under ten.
We then report a seventeen-day unattended field pilot on a working farm in Zimbabwe: 38.7 million frames from six cameras on one embedded device, evaluated against a co-located child-care organisation's daily attendance register.
Software tuning alone improved field detection yield 36-fold, and identity consolidation under a simultaneity veto cut over-reporting from 9.1 times the register to 1.8--3.9 times with zero proven-false merges across 240 verified-distinct pairs.
The pilot also fixes the method's current boundaries by measurement: camera placement dominates model quality, age estimates on small backlit faces are not humanly verifiable, and children under seven remain beyond face-based estimation.
We close with the data-protection and human-in-the-loop safeguards under which such a system should operate.

\end{abstract}

\newpage
\setcounter{page}{1}

\section{Introduction}
\label{sec:introduction}

The most recent joint estimates by the International Labour Organization and UNICEF put 138 million children aged 5--17 in child labour worldwide, 54 million of them in work that directly endangers their health or safety \cite{iloChildLabourGlobal2025,unicefChildLabourGlobal2024}.
The long-term trend is encouraging, with more than 100 million fewer children in child labour than in 2000, yet the target of full elimination by 2025 set under Sustainable Development Goal 8.7 has been missed \cite{iloChildLabourGlobal2025,8.7Alliance87Pathfinder2023}.
Agriculture accounts for roughly 61 per cent of cases, and Sub-Saharan Africa alone carries about 87 million affected children, nearly two thirds of the global total \cite{iloChildLabourGlobal2025}.
In the cocoa belt of Ghana and C\^ote d'Ivoire, the largest survey to date found 45 per cent of children in agricultural households engaged in child labour in cocoa production \cite{chicagoChildLabourCocoa2020}.

Pressure to detect and remediate these situations no longer comes from voluntary commitments alone.
The EU Corporate Sustainability Due Diligence Directive obliges large companies operating in the European market to identify, prevent and account for adverse human-rights impacts, child labour included, throughout their chains of activities \cite{commissionEUCorporateSustainability2024,parliamentEUCorporateSustainability2024}.
The United States maintains a statutory list of goods produced by child or forced labour, currently covering 204 goods from 82 countries \cite{laborListGoodsProduced2024}.
Both instruments build on the ILO minimum-age and worst-forms conventions \cite{organizationILOConvention1381973,organizationILOConvention1821999}.
Companies therefore need monitoring mechanisms whose evidence survives third-party scrutiny.

The instrument the cocoa sector has converged on is the Child Labour Monitoring and Remediation System (CLMRS), a household-visit programme standardised by the International Cocoa Initiative through its Core Criteria and implementation manual \cite{initiativeCLMRSCoreCriteria2025,initiativeCLMRSManual2025}.
CLMRS coverage reached 1.17 million cocoa-farming households by September 2024, about 55 per cent of cocoa-growing households in C\^ote d'Ivoire and Ghana, and more than 970,000 children received preventive or remedial support in 2024 \cite{initiativeTacklingChildLabour2024}.
These systems work, but their measurement layer has well-documented gaps.
The sector-wide effectiveness review, covering twelve projects, over 70,000 farmers and 190,000 child interviews, found identification rates ranging from 0.5 to 26.7 per cent across projects in C\^ote d'Ivoire and from 2.5 to 60.4 per cent in Ghana, and concluded that most systems are unlikely to capture all cases among the households they monitor \cite{initiativeEffectivenessReviewChild2021,iciHowEffectiveAre2023}.
Monitoring visits are episodic, so work performed between visits leaves no trace.
Interview-based measurement is also systematically biased: a recent study that cross-checked parental reports against satellite-verified field observations concluded that parents under-report their children's work by at least 60 per cent \cite{variousMeasuringChildLabor2025}.

This paper asks whether modern computer vision can supply the missing signal: a continuous, presence-based observation channel at workplaces such as processing sites, workshops and farms, feeding candidate sightings of children into the existing CLMRS verification and remediation workflow.
The perceptual task decomposes into locating people and faces in video, estimating the age of each person, and maintaining persistent identities across frames and camera views so that the system counts unique children rather than per-frame hits.
Each sub-task is mature in isolation, but assembling them into a deployable instrument for child-labour monitoring raises requirements that no published system meets simultaneously: child-grade age accuracy rather than a binary adult--child cut, real-time throughput on edge hardware in settings without reliable connectivity, and outputs structured for human verification rather than automated enforcement.

We built and evaluated such a system.
A multi-task detector based on a YOLO26x backbone within the CerberusDet framework \cite{sapkotaYOLO26KeyArchitectural2025,tolstykhCerberusDetUnifiedMultiDataset2024} locates persons and faces in a single forward pass.
A cascaded age estimator pairs the general-purpose MiVOLO~v2 transformer \cite{kuprashevichMiVOLOMultiinputTransformer2023,kuprashevichSpecializationAssessingCapabilities2024} with a child specialist fine-tuned for ages 0--12.
ByteTrack \cite{zhangByteTrackMultiObjectTracking2021} maintains per-camera tracks, and ArcFace \cite{dengArcFaceAdditiveAngular2018} and DINOv2 \cite{oquabDINOv2LearningRobust2023} embeddings merge reappearances into unique persons.
Track-level fusion produces one age, gender and confidence record per person, together with quality flags designed for human triage.
The full pipeline runs above real-time speed on embedded Blackwell-class GPUs after low-precision compilation, and it has been operated in the field: a seventeen-day pilot on a working farm in Zimbabwe, six cameras against a daily attendance register, forms part of the evaluation.

Our contributions are the following.
\begin{enumerate}
  \item \textbf{An end-to-end pipeline for child-labour monitoring.}
  We describe, to our knowledge for the first time in the literature, a complete real-time system that combines multi-task person and face detection, cascaded child-specialist age estimation, tracking, re-identification and CLMRS-compatible reporting, rather than stopping at detection or at a binary age gate (Section~\ref{sec:methods}).
  \item \textbf{A cascade training study with negative results.}
  We report what failed as well as what worked while specialising MiVOLO~v2 for children: catastrophic forgetting under standard fine-tuning rates, a validation-distribution mismatch that masked real progress, and the failure of aggressive oversampling.
  The resulting child specialist reaches a mean absolute error of 1.944 years on children-only validation for ages 0--12 (Sections~\ref{sec:methods} and~\ref{sec:results}).
  \item \textbf{An edge quantisation study on Blackwell-class hardware.}
  FP8 TensorRT compilation of the age cascade yields a 1.77$\times$ speedup at a quality cost of 0.002 years MAE; an FP8 detector engine raises full-pipeline throughput by 18 per cent while halving engine size, at the cost of 2.1 per cent fewer detected children.
  We also document why the FP16 TorchScript path and the FP4 toolchain currently fail for VOLO-family models (Sections~\ref{sec:methods} and~\ref{sec:results}).
  \item \textbf{A multi-level evaluation.}
  We evaluate the detector on 13,537 images against the previous-generation baseline, the age estimator against open-source alternatives on APPA-Real and FairFace-Africa including child subsets, the cascade on 146,735 internal validation samples, and the whole pipeline on 26.8 hours of video, where the new system surfaces 122 per cent more unique child candidates at markedly higher confidence (Section~\ref{sec:results}).
  \item \textbf{A field pilot with register-based ground truth.}
  We report a seventeen-day unattended deployment at a real site in Zimbabwe with children present: 38.7 million frames from six cameras on a single embedded device, evaluated against the co-located child-care organisation's daily attendance register and against a labelling-free false-merge ground truth built from simultaneous-visibility constraints.
  Identity consolidation with a temporal veto cut over-reporting from 9.1 times the register to 1.8--3.9 times at zero proven-false merges, and the pilot located the method's current boundaries, camera geometry, unverifiable age evidence, and the invisibility of the 0--6 band, by measurement rather than conjecture (Sections~\ref{sec:field-methods} and~\ref{sec:results-field}).
\end{enumerate}

Four framing points apply throughout.
First, everything described here is research: the third-party models we build on are used, and our own models and tools were created, solely to study whether vision-based monitoring of this kind is feasible; the system is a research prototype, not a commercial offering.
Second, the end-to-end film corpus consists of documentaries and feature films selected for child-labour-relevant content; it is a demanding proxy without ground-truth annotations, and we treat its numbers as system yield rather than measured recall.
Third, the field pilot has ground truth for presence and merge correctness but not for age, and we identify the site only as far as child protection permits.
Fourth, the system is an assistive instrument.
It proposes candidates with evidence and confidence scores; decisions about children remain with trained CLMRS personnel, a boundary we discuss together with data-protection safeguards in Section~\ref{sec:discussion}.

\section{Background and Related Work}
\label{sec:related}

\subsection{Technology-Based Child-Labour Monitoring}

Existing technological responses to child labour operate mostly at area level.
DIGICHILD, developed by the FAO, estimates child-labour risk on a one-square-kilometre grid from georeferenced poverty, schooling and climate indicators \cite{faoDIGICHILDExploringGeoreferenced2025}.
Satellite-based approaches map risk or detect proxy infrastructure: Sentinel-2 imagery has been used to locate brick kilns across South Asia's kiln belt, a sector strongly associated with bonded and child labour \cite{labSentinelKilnDBScalableBrick2025,labsUNDPBrickKiln2024}, and drone or aerial person detection has been surveyed for humanitarian monitoring \cite{mondalEyeSkyDetection2024}.
These tools answer where to look; they do not observe individual children.

Individual-level systems are rarer and narrower.
\textcite{salmanDeepLearningBased2021} presented the first deep-learning classifier aimed specifically at recognising child labour in images.
\textcite{tahirChildDetectionYOLOv52023} trained a YOLOv5 child detector for CCTV scenarios, and \textcite{variousAgeGroupClassifier2022} demonstrated an embedded adult--child classifier with YOLO-based pre-processing.
\textcite{lehmannFaceSilhouetteAge2022} combined facial and clothed-body silhouette cues to flag minors for a child-protection system.
A commercial service, NoWorKids, applies anthropometric analysis to drone and camera imagery of agricultural fields, but publishes no accuracy figures \cite{noworkidsNoWorKidsIdentifyingChild2024}.
Benchmarking work shows how hard the underlying perception problem is: on a manually annotated image--caption corpus of minors in unconstrained settings, the best of three evaluated detection approaches, one of them a commercial age-estimation product, reached a true-positive rate of only 75.3 per cent \cite{kireevICCWDManuallyAnnotated2025}.
None of these systems combines continuous video operation, child-grade age estimation, identity persistence and edge deployment, which is the combination a workplace monitoring instrument needs.

\subsection{Person and Face Detection}

Single-stage detectors of the YOLO family remain the default choice for real-time deployment \cite{yaseenWhatYOLOv8InDepth2024,khanamYOLOv11OverviewKey2024}.
The recent YOLO26 generation removes distribution focal loss, adopts natively NMS-free inference and introduces progressive loss balancing and small-target-aware label assignment, with reported gains concentrated exactly where surveillance workloads hurt: small objects and dense scenes \cite{sapkotaYOLO26KeyArchitectural2025,sapkotaUltralyticsYOLOEvolution2025}.
Transformer detectors such as DINO and RT-DETR are competitive in accuracy \cite{zhangDINODETRImproved2023,zhaoDETRsBeatYOLOs2024}, but their latency profiles on embedded GPUs still favour YOLO-class models for our setting.
Detecting persons and faces jointly is a multi-task problem; CerberusDet shows that a shared backbone with per-task necks and heads matches separately trained detectors while sharing most of the computation \cite{tolstykhCerberusDetUnifiedMultiDataset2024}.
Child-specific detection studies confirm that children are harder targets than adults, motivating dedicated tuning and evaluation \cite{tranEnhancingYOLOv11nReliable2026,arxiv:2506.13445OvercomingOcclusionsWild2025}.

\subsection{Facial Age Estimation}

Age estimation from faces has progressed from expectation-based CNN regression \cite{rotheDEXDeepEXpectation2015} through ordinal and rank-consistent formulations \cite{caoRankConsistentOrdinal2019,caoCORALRankConsistent2020}, distribution-aware losses \cite{panMeanVarianceLossDeep2018} and compact architectures \cite{yangSSRNetCompactSoft2018,zhangC3AEExploringLimits2019} to transformer models.
MiVOLO processes a face crop and a body crop jointly, which preserves an age signal when the face is small, averted or occluded, and holds state-of-the-art results across public benchmarks \cite{kuprashevichMiVOLOMultiinputTransformer2023,kuprashevichSpecializationAssessingCapabilities2024}.
Systematic evaluations caution that headline benchmark numbers transfer poorly across datasets and demographic groups \cite{paplhamCallReflectEvaluation2023,papadopoulosBenchmarkingDeepLearning2024}.
Two failure modes matter most for child protection.
Models trained on adult-dominated corpora estimate children poorly, a data gap documented across benchmarks \cite{karkkainenFairFaceFaceAttribute2021}, and accuracy collapses precisely in the adolescent band where legal working-age thresholds sit \cite{gaulUnderageDetectionMultiTask2025,dantchevaVec2UAgeEnhancingUnderage2021}.
Under-age detection research therefore evaluates threshold-level error rates, such as the false-adult rate at 16 or 18, rather than MAE alone \cite{variousUnderageDetectionMultiTask2025}.
Demographic bias is a further documented risk \cite{variousEthnicRepresentationMatters2024,cliffordTwoSourcesBias2018}.
Our cascade design responds to the first failure mode with a child-specialist model; the second remains a limitation we quantify and discuss.

\subsection{Tracking, Re-Identification and Edge Inference}

Tracking-by-detection with ByteTrack retains low-confidence boxes during association, which preserves partially occluded people, the common case in workplaces \cite{zhangByteTrackMultiObjectTracking2021}; appearance-augmented variants such as BoT-SORT, StrongSORT and OC-SORT occupy nearby design points \cite{aharonBoTSORTRobustAssociations2022,duStrongSORTMakeDeepSORT2022,caoObservationCentricSORTRethinking2022,variousMultiObjectTrackingReview2025}.
For merging track fragments into unique persons, ArcFace face embeddings remain the standard \cite{dengArcFaceAdditiveAngular2018}, and self-supervised DINOv2 features provide a robust body-appearance signal without task-specific training \cite{oquabDINOv2LearningRobust2023}.
On the deployment side, FP8 formats with hardware support on recent NVIDIA GPUs offer near-FP16 accuracy at reduced memory bandwidth \cite{kuzminFP8QuantizationPower2024,vanbaalenFP8INT8Efficient2023,zhangFP8BERTPostTrainingQuantization2023}, and TensorRT is the usual vehicle for such engines \cite{kimTensorRTBasedFrameworkOptimization2022,nvidiaNVIDIATensorRTModel2025}.
Post-training quantisation of vision transformers has its own literature, because activation distributions in attention blocks quantise badly without reparameterisation \cite{yuanPTQ4ViTPostTrainingQuantization2022,liRepQViTScaleReparameterization2023,wuMPTQViTMixedPrecisionPostTraining2024}; YOLO-family detectors are comparatively well behaved under low-bit conversion \cite{variousQuantizingYOLOv7Comprehensive2024,variousYOLOv8INT8Quantization2025}.
What the literature lacks is an account of these techniques composed into one child-monitoring system and measured end to end, which is the gap this paper addresses.

\section{System Design and Methods}
\label{sec:methods}

\subsection{Design Requirements}
\label{sec:requirements}

Four requirements shaped the design.
First, the system must run continuously at or above real-time speed on edge hardware, because target sites, such as farms, processing stations and workshops in rural producing regions, cannot rely on connectivity to cloud services.
Second, age must be estimated as a continuous value with dedicated accuracy in childhood, not as a binary adult--child decision: remediation priority differs sharply between a five-year-old and a fifteen-year-old.
The internal technical specification required an overall MAE of at most 3.0 years and at most 2.5 years for ages 4--16.
Third, the unit of reporting must be the unique person, which requires tracking within a camera and re-identification across gaps and viewpoints; per-frame detections would inflate counts and be useless for case management.
Fourth, every output must be reviewable: a person record carries its evidence (best-shot crops, per-frame trajectory, confidence and quality flags) so that trained CLMRS personnel can verify or dismiss it.
The system proposes; it never adjudicates.

\subsection{Pipeline Overview}
\label{sec:pipeline}

\begin{figure}[htbp]
  \centering
  \begin{tikzpicture}[
      node distance=0.55cm and 0.55cm,
      stage/.style={draw, rounded corners=2pt, align=center, font=\small, inner sep=5pt, minimum height=0.95cm},
      arrow/.style={-{Latex[length=2.2mm]}, thick}
    ]
    \node[stage] (video) {Video\\stream};
    \node[stage, right=of video] (det) {Multi-task detector\\YOLO26x + CerberusDet\\(person + face)};
    \node[stage, right=of det] (age) {Cascaded age/gender\\MiVOLO~v2 general\\+ child specialist};
    \node[stage, below=of age] (track) {Tracking\\ByteTrack};
    \node[stage, left=of track] (reid) {Re-identification\\ArcFace + DINOv2};
    \node[stage, left=of reid] (fusion) {Track-level fusion,\\deduplication,\\quality flags};
    \node[stage, below=of reid] (daily) {Daily aggregation,\\identity consolidation\\(field deployments)};
    \node[stage, below=of fusion] (report) {Person records for\\human review (CLMRS)};
    \draw[arrow] (video) -- (det);
    \draw[arrow] (det) -- (age);
    \draw[arrow] (age) -- (track);
    \draw[arrow] (track) -- (reid);
    \draw[arrow] (reid) -- (fusion);
    \draw[arrow] (fusion) -- (report);
    \draw[arrow] (fusion) -- (daily);
    \draw[arrow] (daily) -- (report);
  \end{tikzpicture}
  \caption{Processing pipeline. Detections feed the age cascade; tracking and re-identification merge frame-level evidence into unique persons; fusion produces one reviewable record per person. Field deployments add a daily aggregation and identity-consolidation layer (Section~\ref{sec:field-methods}).}
  \label{fig:pipeline}
\end{figure}
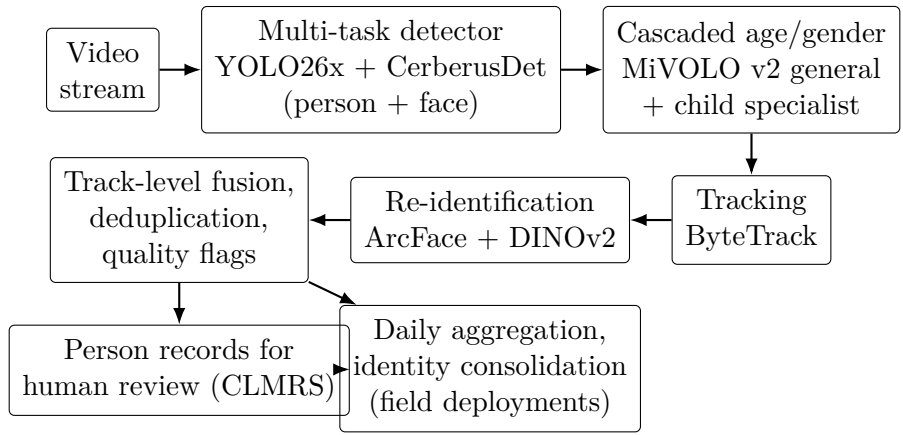

Figure~\ref{fig:pipeline} shows the pipeline.
Continuous RTSP camera streams are processed in ten-minute segments by one persistent worker per camera.
A multi-task detector locates persons and faces in each frame.
For every person with an associated face, the age module receives a six-channel input, a face crop and a body crop of three channels each, and returns age and gender.
ByteTrack associates detections over time into tracks; ArcFace face embeddings and DINOv2 body embeddings merge tracks that belong to the same individual, within and across reappearances, at a cosine-similarity threshold of 0.70.
A fusion stage aggregates per-frame estimates into a per-person age and gender with a fusion score in $[0,1]$, discards tracks visible for less than 3.5 seconds, and attaches quality flags (short visibility, high age variance, low gender confidence and nine further flag types) that support human triage.
A person whose fused age falls below the child threshold is recorded as a child candidate.
In field deployments a further layer aggregates encounters across the day and consolidates identities before anything is filed; the laboratory studies of Sections~\ref{sec:results-detector}--\ref{sec:results-pipeline} run without it, and Section~\ref{sec:field-methods} describes it in full.
An earlier pipeline generation (v5.0) additionally ran pose estimation and a voice-based age branch; both were removed because their contribution did not justify their cost on edge hardware.

\subsection{Multi-Task Person and Face Detector}
\label{sec:detector}

\subsubsection{Architecture}

The detector combines a YOLO26x backbone with the CerberusDet multi-task architecture \cite{sapkotaYOLO26KeyArchitectural2025,tolstykhCerberusDetUnifiedMultiDataset2024}.
The backbone (C3k2 and C2PSA blocks with SPPF, 28.8M parameters) is shared; the neck branches after its second block into task-specific tails, so person detection and face detection each own a single-class, anchor-free detection head over feature levels P3--P5 (strides 8, 16, 32).
The full model has 101.4M parameters and 366 GFLOPs at $640\times640$ input.
Joint detection in one forward pass avoids running two detectors, and per-task heads avoid the capacity competition that a single two-class head exhibits between large bodies and small faces.

\subsubsection{Training Objective}

Each task $t \in \{\text{person}, \text{face}\}$ is trained with the composite detection loss
\begin{equation}
  \label{eq:det-loss}
  \mathcal{L}_t \;=\; \lambda_{\mathrm{box}}\,\mathcal{L}_{\mathrm{CIoU}} \;+\; \lambda_{\mathrm{cls}}\,\mathcal{L}_{\mathrm{BCE}} \;+\; \lambda_{\mathrm{dfl}}\,\mathcal{L}_{\mathrm{DFL}},
\end{equation}
with $(\lambda_{\mathrm{box}}, \lambda_{\mathrm{cls}}, \lambda_{\mathrm{dfl}}) = (7.5, 0.5, 1.5)$ for both tasks.
Positive anchors are chosen by task-aligned assignment: for every ground-truth box each candidate anchor receives the alignment score
\begin{equation}
  \label{eq:tal}
  m \;=\; s^{\alpha}\,\mathrm{IoU}(\hat{b}, b)^{\beta},
  \qquad \alpha = 0.5,\; \beta = 6.0,
\end{equation}
where $s$ is the predicted classification score and $\hat{b}$ the decoded box; the top ten anchors per ground truth become positives, and the classification branch is trained with binary cross-entropy against the normalised alignment scores, so localisation quality supervises classification.
Box sides are regressed as discrete distributions over $R{+}1$ bins: the decoded distance is the expectation $\hat{d} = \sum_{j=0}^{R} j\,\mathrm{softmax}(z)_j$, and the distribution focal loss interpolates cross-entropy between the two integer bins bracketing the target distance $d$,
\begin{equation}
  \label{eq:dfl}
  \mathcal{L}_{\mathrm{DFL}} \;=\; -\big( (d_{+}-d)\,\log p_{d_{-}} + (d-d_{-})\,\log p_{d_{+}} \big),
\end{equation}
with $d_{-} = \lfloor d \rfloor$ and $d_{+} = d_{-}+1$.
The shared backbone is updated with gradients averaged across the alternating task batches, which is the CerberusDet parameter-sharing scheme \cite{tolstykhCerberusDetUnifiedMultiDataset2024}.

\subsubsection{Training Configuration}

We fine-tuned the detector from a YOLO26x checkpoint on public datasets converted to a unified two-task format: COCO Person, Objects365 Person and CrowdHuman for the person task (108,470 training and 10,315 validation images), and WiderFace, CrowdHuman heads, Hollywood Heads and FDDB, complemented by COCO pseudo-face labels generated with the MiVOLO reference detector, for the face task (63,130 training and 8,174 validation images).
Training ran on four NVIDIA H200 GPUs with distributed data parallelism, synchronised batch normalisation and mixed precision, batch size 192 at $640\times640$, SGD with an initial learning rate of 0.00309 under a one-cycle schedule, momentum 0.952 and weight decay of $3.7\times10^{-4}$.
Augmentation hyperparameters were obtained by evolutionary search and include mosaic (probability 1.0), mixup (0.285), aggressive scale jitter (0.846) and mild blur and greyscale perturbations (Appendix~\ref{app:detector-config}).
Early stopping with a patience of 50 epochs selected epoch 115 of 166 after 71.7 hours of training.
On the held-out validation sets the selected checkpoint reaches a person mAP@0.5 of 0.872 (mAP@0.5:0.95 of 0.607) and a face mAP@0.5 of 0.755 (mAP@0.5:0.95 of 0.448).
Section~\ref{sec:results-detector} reports the controlled comparison against the previous-generation baseline detector on an independent benchmark.

\subsection{Cascaded Age and Gender Estimation}
\label{sec:cascade}

\subsubsection{Architecture}
\label{sec:mivolo-arch}

Age and gender estimation builds on MiVOLO~v2 \cite{kuprashevichMiVOLOMultiinputTransformer2023,kuprashevichSpecializationAssessingCapabilities2024}, a VOLO-D1 vision transformer operating at $384\times384$ whose first stage uses outlook attention, a mechanism that aggregates each spatial position's local neighbourhood with learned per-position attention weights before the usual self-attention stages.
The model consumes a six-channel input, the face crop and the person crop concatenated channel-wise, and its patch embedding is modified to process the two streams in parallel stems fused by a bidirectional cross-attention block.
Denoting the face-stream and person-stream feature maps $X_f$ and $X_p$, $1\times1$ convolutions produce per-stream queries, keys and values, and each stream attends to the other:
\begin{equation}
  \label{eq:crossattn}
  O_f = \mathrm{softmax}\!\big(q_f k_p^{\top}/\sqrt{d} + P\big)\,v_p,
  \qquad
  O_p = \mathrm{softmax}\!\big(q_p k_f^{\top}/\sqrt{d} + P\big)\,v_f,
\end{equation}
where $P$ is a relative positional embedding; the concatenated outputs pass through layer normalisation and a GELU MLP, $Y = \mathrm{MLP}(\mathrm{LN}([O_f; O_p]))$.
Face queries thus read body context (posture, clothing, scale) and body queries read facial detail, which is what lets the model degrade gracefully when one modality is small, averted or missing.
The head outputs the triple $(g_m, g_f, a_{\mathrm{norm}})$, gender logits and a normalised age, and the age is denormalised as
\begin{equation}
  \label{eq:agedenorm}
  a \;=\; a_{\mathrm{norm}}\,(a_{\max} - a_{\min}) + \bar{a},
\end{equation}
with $(a_{\min}, a_{\max}, \bar{a}) = (0, 122, 61.0)$ in the deployed cascade metadata.

\subsubsection{Cascade Structure}

Rather than force one model to serve all ages, we run a two-model cascade of the general model $G$ and a child specialist $C$, both full MiVOLO~v2 instances sharing the input format:
\begin{equation}
  \label{eq:cascade}
  a \;=\;
  \begin{cases}
    a_C(x), & a_G(x) < \tau_{\mathrm{route}},\\[2pt]
    a_G(x), & \text{otherwise},
  \end{cases}
\end{equation}
with gender always taken from $G$.
The routing threshold is $\tau_{\mathrm{route}} = 20$ years in the laboratory configuration, chosen conservatively so that few genuine children bypass the specialist.
Field experience revised this value: on small, backlit faces the FP8 general model collapses towards its dataset prior of roughly 27.8 years, so children never crossed the threshold and never reached the specialist.
The deployed pilot therefore runs $\tau_{\mathrm{route}} = 32$, validated on site: the specialist's own prior on uninformative faces is about 22 years, safely above the child-classification threshold, so widening the route recovers children without flagging adults.
Section~\ref{sec:results-cascade} quantifies the cost of threshold choice for adolescents, and Section~\ref{sec:discussion} discusses it further.

\subsubsection{Training Protocol and Negative Results}
\label{sec:training-protocol}

We report the training path in full, including three failed strategies, because the failures shaped the final design and are, in our view, of practical value to groups attempting similar specialisation.
The available pool comprised roughly 1.13 million age-labelled face images across fourteen public datasets (IMDB-Wiki, AFAD, CACD, FairFace, MORPH-2, MegaAge-Asian, FairFace-Africa, All-Age-Faces, UTKFace, APPA-Real, Adience, AgeDB, a partial copy of LAGENDA, and FG-NET).

An initial six-phase plan with a learning rate of $5\times10^{-5}$, teacher pseudo-labelling of unlabelled detection datasets and self-distillation degraded the model monotonically.
Three lessons emerged.
A learning rate conventional for from-scratch training destroys a well-trained initialisation; the working range proved to be $1\times10^{-6}$ to $5\times10^{-6}$, roughly 25 times lower than planned.
Pseudo-labels from a teacher with an MAE of 3.65 cap the student near the teacher's error and drown the signal of clean labels.
Self-distillation, with the same model as teacher and student, is circular and gained nothing.

A conservative re-plan (frozen-backbone warm-up, then fine-tuning on eight clean datasets at $2\times10^{-6}$, then on thirteen datasets at $1\times10^{-6}$) improved the general model's validation MAE from about 5.0 to 3.936.
Two further attempts to make this single model good at children failed instructively.
Training on children-only data appeared to degrade the model until we found that the validation loader ignored the children-only filter: the model was improving on children while being scored on all ages.
After the mismatch was fixed, specialisation worked.
A weighted-sampling alternative, oversampling children tenfold within all-ages training, degraded adults faster than it improved children and was abandoned.

The final child specialist was fine-tuned from the best general checkpoint on children-only samples (119,233 training, 19,975 validation) in three phases: a frozen-backbone warm-up, 40 epochs on ages 0--18 at $5\times10^{-6}$, and a 15-epoch focus phase on ages 0--12 at $1\times10^{-6}$ (Appendix~\ref{app:age-phases}).
It reaches an MAE of 2.291 years for ages 0--18 and 1.944 years for ages 0--12, with CS@5 of 93.9 per cent, on children-only validation.
A companion result concerns the general slot of the cascade: on public benchmarks the fine-tuned general model lost to the unmodified MiVOLO~v2 by 8 to 18 per cent MAE (catastrophic forgetting concentrated in the under-represented age extremes), so the reference configuration pairs the original general model with our specialist.
Section~\ref{sec:results-age} presents both configurations.

\subsection{Tracking, Re-Identification and Fusion}
\label{sec:tracking}

ByteTrack performs tracking-by-detection and keeps low-confidence boxes in the association step, which preserves partially occluded persons \cite{zhangByteTrackMultiObjectTracking2021}.
Track fragments are merged into persons using ArcFace embeddings $e \in \mathbb{R}^{512}$ of the best face crops \cite{dengArcFaceAdditiveAngular2018} and DINOv2 ViT-S/14 embeddings of body crops \cite{oquabDINOv2LearningRobust2023}, thresholded at cosine similarity $\cos(e_i, e_j) \geq 0.70$.
Along a track the face embedding is maintained as an exponential moving average, $e_t = (1-\eta)\,e_{t-1} + \eta\,e_{\mathrm{obs}}$; a direct comparison against exemplar-set representations on a 6,862-crop corpus from the field pilot confirmed the averaged representation superior at every candidate set size.
On the film corpus of Section~\ref{sec:results-pipeline} roughly 60 per cent of raw candidates merge into existing persons, which shows how misleading per-track counting would be.

Fusion converts a track's evidence into a classification.
Only face-backed age estimates enter the age history: samples computed with zeroed face channels are tagged and excluded, because without a face the model returns its dataset prior rather than information (a field measurement puts that prior at 22.4 years in 89 per cent of body-only cases).
From the face-backed history $A$ the robust age is the 25th percentile, $\hat{a} = P_{25}(A)$, a deliberately low statistic because small and poorly lit faces bias estimates upwards, so a low percentile tracks children's true age better while adults with many samples remain far above the threshold.
The child probability is a sigmoid around the age threshold $\tau_{\mathrm{child}}$ with a hard veto above $a_{\max}$:
\begin{equation}
  \label{eq:fusion}
  p \;=\;
  \begin{cases}
    0, & \hat{a} > a_{\max},\\[2pt]
    \sigma\big(k\,(\tau_{\mathrm{child}} - \hat{a})\big), & \text{otherwise},
  \end{cases}
  \qquad k = 0.5,
\end{equation}
and a person is classified as a child when $p \geq 0.5$.
The laboratory configuration uses $\tau_{\mathrm{child}} = 16$ and $a_{\max} = 19$; the field pilot runs $\tau_{\mathrm{child}} = 18$ and $a_{\max} = 21$, and additionally requires $p \geq 0.55$ before an evidence package may be filed.
The fusion score reported throughout this paper is this probability $p$.
Records are tiered for review (high $\geq 0.8$, medium 0.6--0.8, low otherwise).

\subsection{Field Aggregation and Identity Consolidation}
\label{sec:field-methods}

Continuous multi-camera operation raises a problem the film corpus does not: the same child appears dozens of times per day, across segments and cameras, and naive per-segment filing multiplies one child into many reports.
The field deployment therefore adds an aggregation layer, developed and validated during the pilot of Section~\ref{sec:results-field}.

\paragraph{Daily windows.}
Workers no longer file per segment; they register \emph{encounters} in a shared identity registry (a transactional SQLite store, about one millisecond per segment), and a cutoff process at local midnight assembles the day's evidence.
Per-encounter matching joins a new encounter to an existing identity when face similarity clears 0.75 across cameras or body similarity clears 0.72 within the window, subject to an age gate of five years.
The face threshold was raised from an initial 0.62 after a measurement on 6,862 diagnostic face crops (289 same-person and 675 different-person pairs) located the old value inside the different-person similarity distribution (median 0.542, 99th percentile 0.715); the empirical threshold at a one per cent false-merge rate is 0.760, and the deployed 0.75 raised labelled merge precision from 0.50 to 0.67.

\paragraph{Consolidation with a simultaneity veto.}
Per-encounter matching is sequential and never revisits pairs of already-formed identities, so one child can still fragment into several identities.
At window close a union-find pass merges identity groups $G_i, G_j$ when their best pairwise face similarity reaches 0.66, or, where no usable face comparison exists between the groups, when body similarity reaches 0.61 (a sub-threshold face comparison vetoes the body path), subject to a hard \emph{time guard}:
\begin{equation}
  \label{eq:timeguard}
  \mathrm{merge}(G_i, G_j) \text{ is forbidden if } \exists\, e \in G_i,\, e' \in G_j:\; \big[t_s(e), t_e(e)\big] \cap \big[t_s(e'), t_e(e')\big] \neq \emptyset,
\end{equation}
that is, two groups may not merge if any of their encounters were visible simultaneously, because one person cannot be two boxes at the same moment.
The check is group-to-group rather than pair-to-pair, otherwise transitive unions ($A{+}B$, then $B{+}C$) drag a forbidden $A{+}C$ merge through.
The veto is what makes aggressive similarity thresholds safe, and Section~\ref{sec:results-field} shows what happens without it.
Two further guards close specific defect classes found on site: an identity may absorb at most one encounter per segment (two children on screen together were otherwise occasionally merged, hiding one child inside another's record), and an accuracy gate re-validates the child classification against the \emph{aggregated} age before filing, because aggregation initially recomputed the mean age while copying the segment-level child flag, once producing a filed report for a person re-estimated at 29.3 years.

\paragraph{Face rescue.}
A second-pass face detector (the InsightFace SCRFD family) re-scans person crops for which the primary detector returned no face, at reduced resolution and bounded count; it recovers two to six per cent additional faces, confirming that most faceless tracks genuinely show no visible face.

\subsection{Edge Deployment and Quantisation}
\label{sec:quantisation}

The deployment targets are embedded Blackwell-class machines: an NVIDIA DGX Spark (GB10, compute capability 12.1) for the laboratory studies and an NVIDIA Jetson AGX Thor (compute capability 11.0) in the field, both running the pipeline as containerised services (PyTorch 2.10, Torch-TensorRT, TensorRT 10.x).
Both MiVOLO models are compiled to TensorRT engines with native FP8 (E4M3) precision through the Torch-TensorRT dynamo path \cite{nvidiaNVIDIATensorRTModel2025,kuzminFP8QuantizationPower2024}.
FP8 quantisation maps a tensor $x$ to $\hat{x} = s \cdot \mathrm{clip}\!\big(\mathrm{round}(x/s), -q_{\max}, q_{\max}\big)$ with per-tensor scales $s$ chosen on calibration data; the E4M3 format spans $\pm 448$ with three mantissa bits, and its hardware support on Blackwell GPUs is what converts the reduced precision into memory-bandwidth savings.
VOLO's dynamic operations (\texttt{col2im}, \texttt{unfold}, \texttt{cat}, \texttt{slice}) do not convert cleanly, so these operations run in PyTorch fallback via \texttt{torch\_executed\_ops}; the engines shrink from 110~MB to 93~MB per model, and the maximum output deviation from FP32 on the calibration sample was $5.6\times10^{-4}$ on GB10 and between $10^{-4}$ and $10^{-3}$ on Thor, where the engines were rebuilt on device because TensorRT engines do not port across GPU architectures.

Three negative toolchain results are worth recording.
FP16 engines compiled through the older TorchScript IR fail at runtime for VOLO, because the converter splits the graph into sub-engines and routes a scalar where a shaped tensor is expected; the dynamo path does not exhibit the problem.
FP4 (NVFP4) quantisation could not be realised on either platform: ModelOpt's fake-quantisation tensors are not traceable by \texttt{torch.export}, and on Thor (TensorRT 10.14, ModelOpt 0.39) requesting the FP4 builder flag without explicit quantise--dequantise nodes silently produces FP16 kernels, a configuration that looks like FP4 and is not.
The detector ships as TensorRT engines in two variants, FP16 (197--208~MB) and an output-quantised FP8 build (104~MB): on this model FP16 costs 0.2 to 0.4 per cent accuracy against the FP32 ONNX reference while FP8 costs 3 to 5 per cent, so the field pilot deploys the FP16 engine and the FP8 variant serves as the throughput ablation of Section~\ref{sec:results-quant}.

\subsection{Evaluation Methodology}
\label{sec:eval-method}

We evaluate at five levels, using data disjoint from training wherever accuracy is scored.

\paragraph{Detector benchmark.}
The current and previous detector generations are compared on a combined validation set of 13,537 images with 156,226 ground-truth boxes (117,114 person, 39,112 face) drawn from COCO Person (2,693 images), WiderFace (3,222), CrowdHuman (4,369) and Objects365 Person (3,253).
Both models run at $640\times640$ in FP16 on the same GB10 GPU, with a confidence floor of 0.001 and NMS IoU of 0.7, and are scored with pycocotools against identical annotations.
Two caveats hold: the baseline ran batched inference while the new model ran per-image, and part of the face ground truth derives from CrowdHuman head boxes, which depresses absolute face mAP for both models equally.

\paragraph{Age benchmarks against open alternatives.}
On identical face crops from the APPA-Real validation set (1,500 images, real-age labels) and a FairFace-Africa validation subset (6,165 images), we compare four configurations of our system against two widely used open-source stacks, InsightFace (buffalo\_l, genderage) and DeepFace's age module.
Metrics are MAE, median absolute error, CS@5 and CS@10 (the share of predictions within five and ten years), reported for the full sets, for child subsets (ages $\leq 16$: 261 and 2,736 faces respectively) and per age bucket.

\paragraph{Internal cascade evaluation.}
The cascade and its ablations are evaluated on 146,735 held-out validation samples pooled from fourteen datasets, with per-age-group MAE, CS@5 and the share of samples routed to the child model.

\paragraph{End-to-end film study.}
Pipeline generations v5.0 and v6.1 process the same corpus of nineteen videos (26.8 hours, 482,486 frames): documentaries on child labour in cocoa growing and mining, feature films with child protagonists, and control material with few or no children.
The corpus is a stress proxy chosen for difficult cinematography and child-rich scenes; it carries no ground-truth identity or age annotations, so we report system yield (unique child candidates, confidence distributions, quality flags, throughput), explicitly not recall or precision.
The FP16-versus-FP8 detector ablation reuses the same corpus with every other component held fixed.

\paragraph{Field pilot.}
The system ran for seventeen days (28 July to 13 August 2026, with the pilot extended beyond the snapshot date) on a working farm in Zimbabwe, observing the premises through six fixed 1080p RTSP cameras processed at five frames per second each by a single Jetson AGX Thor on site.
The children present belong to a registered residential child-care organisation located on the same site; the deployment operated under agreement with both the site owner and that organisation, and we deliberately do not name either, for the children's protection.
Ground truth takes four forms, none requiring image annotation of children: the organisation's daily attendance register (16 to 23 children per day across fourteen covered windows, composition changing daily); \emph{impossible pairs}, encounters whose visibility intervals overlap on the same camera and therefore show different people by construction (240 proven-different pairs over four days, audited individually); ByteTrack track identity as a labelling primitive for re-identification measurements (a 6,862-crop corpus); and 58 hand-labelled evidence items.
Field metrics are accordingly yield against the register (an over-reporting factor), conditional recall on segments where a child's face was visible, false merges against impossible pairs, and throughput and stability of the deployment itself.
Candidate reports were filed into the partner's case-management system throughout, under the study protocol agreed with the partner, so the pilot exercises the complete reporting path while remaining a feasibility study.

\section{Results}
\label{sec:results}

\subsection{Detector Benchmark}
\label{sec:results-detector}

Table~\ref{tab:detector} compares the new multi-task detector against the previous-generation baseline, a YOLOv8x fine-tuned for the same two classes, on the combined 13,537-image benchmark described in Section~\ref{sec:eval-method}.

\begin{table}[htbp]
  \centering
  \caption{Detector comparison on 13,537 validation images (pycocotools COCO evaluation, identical ground truth). AR@100 denotes average recall at 100 detections per image.}
  \label{tab:detector}
  \begin{tabular}{llccc}
    \toprule
    Class & Metric & YOLOv8x-pf & YOLO26x-pf & $\Delta$ rel. \\
    \midrule
    all    & mAP@0.5      & 0.259 & \textbf{0.415} & $+60.3\%$ \\
    all    & mAP@0.5:0.95 & 0.131 & \textbf{0.257} & $+96.9\%$ \\
    all    & AR@100       & 0.339 & \textbf{0.513} & $+51.3\%$ \\
    \midrule
    person & mAP@0.5      & 0.390 & \textbf{0.683} & $+75.0\%$ \\
    person & mAP@0.75     & 0.162 & \textbf{0.460} & $+184.6\%$ \\
    person & mAP@0.5:0.95 & 0.191 & \textbf{0.438} & $+128.8\%$ \\
    person & AR@100       & 0.380 & \textbf{0.622} & $+63.8\%$ \\
    \midrule
    face   & mAP@0.5      & 0.127 & \textbf{0.146} & $+15.2\%$ \\
    face   & mAP@0.5:0.95 & 0.070 & \textbf{0.077} & $+9.1\%$ \\
    face   & AR@100       & 0.298 & \textbf{0.404} & $+35.5\%$ \\
    \midrule
    ---    & FPS          & 51.4  & \textbf{76.1}  & $+47.9\%$ \\
    \bottomrule
  \end{tabular}
\end{table}

Person detection improves radically: mAP@0.5 rises from 0.390 to 0.683, strict-localisation mAP@0.75 nearly triples, and average recall gains 24.2 percentage points.
Face detection gains are more modest in mAP but substantial in recall (AR@100 from 0.298 to 0.404), and the new model produces 46 per cent more detections overall (1,531,954 against 1,049,992).
Recall is the metric that matters most here: a person or face that is never detected can never be age-estimated, so every recall point removes a blind spot.
Absolute face numbers are depressed for both models by the CrowdHuman head-box annotations in the ground truth, as noted in Section~\ref{sec:eval-method}, but the comparison between models is unaffected.
Despite running per-image rather than batched, the new detector is also 47.9 per cent faster on the same GPU.

\subsection{Age Estimation Against Open Alternatives}
\label{sec:results-age}

Table~\ref{tab:age-bench} reports age-estimation accuracy on identical face crops for our cascade (the FP8 reference build and its FP32 counterpart), the ablation with the fine-tuned general model, the unmodified single MiVOLO~v2, and two widely deployed open-source stacks.

\begin{table}[htbp]
  \centering
  \caption{Age estimation on APPA-Real (n=1{,}500) and FairFace-Africa (n=6{,}165). CS@5/CS@10: share of predictions within 5/10 years. Best value per column in bold.}
  \label{tab:age-bench}
  \begin{tabular}{lcccccc}
    \toprule
    & \multicolumn{3}{c}{APPA-Real} & \multicolumn{3}{c}{FairFace-Africa} \\
    \cmidrule(lr){2-4} \cmidrule(lr){5-7}
    Model & MAE & CS@5 & CS@10 & MAE & CS@5 & CS@10 \\
    \midrule
    Cascade, FP8 TRT (reference) & 4.76 & 62.6\% & 90.4\% & 5.26 & 60.4\% & 86.1\% \\
    Cascade, FP32                 & 4.75 & 62.7\% & 90.5\% & 5.26 & 60.5\% & 86.2\% \\
    Cascade, fine-tuned general   & 5.19 & 57.9\% & 87.9\% & 6.38 & 51.7\% & 79.4\% \\
    MiVOLO~v2, single             & \textbf{4.63} & \textbf{64.3\%} & \textbf{90.9\%} & \textbf{4.90} & \textbf{63.8\%} & \textbf{87.7\%} \\
    InsightFace (buffalo\_l)      & 13.67 & 22.1\% & 44.5\% & 13.51 & 23.4\% & 43.6\% \\
    DeepFace (age)                & 10.33 & 35.1\% & 59.8\% & 13.83 & 23.2\% & 42.3\% \\
    \bottomrule
  \end{tabular}
\end{table}

\begin{table}[htbp]
  \centering
  \caption{Child subsets (age $\leq 16$) of the same benchmarks: APPA-Real (n=261) and FairFace-Africa (n=2{,}736).}
  \label{tab:age-children}
  \begin{tabular}{lcccc}
    \toprule
    & \multicolumn{2}{c}{APPA-Real} & \multicolumn{2}{c}{FairFace-Africa} \\
    \cmidrule(lr){2-3} \cmidrule(lr){4-5}
    Model & child MAE & child CS@5 & child MAE & child CS@5 \\
    \midrule
    Cascade, FP8 TRT (reference) & 4.74 & 62.8\% & 4.64 & 65.5\% \\
    Cascade, FP32                 & 4.75 & 62.8\% & 4.64 & 65.6\% \\
    Cascade, fine-tuned general   & 5.10 & 61.3\% & 6.80 & 51.3\% \\
    MiVOLO~v2, single             & \textbf{4.72} & \textbf{64.4\%} & \textbf{4.17} & \textbf{73.0\%} \\
    InsightFace (buffalo\_l)      & 22.93 & 3.4\%  & 18.74 & 5.6\% \\
    DeepFace (age)                & 18.67 & 0.8\%  & 20.17 & 0.1\% \\
    \bottomrule
  \end{tabular}
\end{table}

Three observations follow.
First, the open-source stacks that practitioners might reach for by default are unusable on children.
On the child subsets (Table~\ref{tab:age-children}) InsightFace and DeepFace show MAEs between 18.7 and 22.9 years, and their per-bucket means reveal the mechanism: children aged 5--12 receive mean predictions of 27 to 35 years.
Both systems systematically turn children into adults, the worst possible failure mode for a protection instrument.
Second, the fine-tuned general model is worse than the original everywhere, by 8.4 per cent MAE on APPA-Real and 18.3 per cent on FairFace-Africa, which is the catastrophic-forgetting result of Section~\ref{sec:training-protocol} confirmed on independent data.
Third, the unmodified single MiVOLO~v2 slightly outperforms the cascade on these two benchmarks, including their child subsets.
This deserves an honest reading, which the per-bucket breakdown provides.

\begin{table}[htbp]
  \centering
  \caption{Per-bucket MAE of the reference cascade (FP8) and the single MiVOLO~v2 on FairFace-Africa. Buckets with few samples on APPA-Real behave analogously.}
  \label{tab:age-buckets}
  \begin{tabular}{lccc}
    \toprule
    Age bucket & n & Cascade FP8 & MiVOLO~v2 single \\
    \midrule
    0--4   & 199   & \textbf{2.08} & 2.66 \\
    5--12  & 1{,}356 & \textbf{2.63} & 2.94 \\
    13--19 & 1{,}181 & 7.37 & \textbf{5.83} \\
    20--39 & 2{,}378 & 5.60 & \textbf{5.21} \\
    40--59 & 880   & 6.08 & \textbf{6.07} \\
    60+    & 171   & 6.28 & \textbf{6.28} \\
    \bottomrule
  \end{tabular}
\end{table}

Table~\ref{tab:age-buckets} shows that the cascade wins where the specialist actually operates, ages 0--4 (2.08 versus 2.66) and 5--12 (2.63 versus 2.94), and loses in the 13--19 bucket, where the routing threshold of 20 sends adolescents to a model trained for ages 0--12.
The child subsets of Table~\ref{tab:age-children} are dominated by that adolescent band, which explains the aggregate ordering.
For CLMRS purposes the youngest bands carry the highest protection weight, and there the cascade is the stronger configuration; the adolescent band is a limitation both configurations share, and we return to it in Sections~\ref{sec:results-cascade} and~\ref{sec:discussion}.

\subsection{Internal Cascade Evaluation}
\label{sec:results-cascade}

Table~\ref{tab:cascade-internal} reports the cascade on 146,735 held-out samples pooled from fourteen datasets, together with the share of samples the threshold routes to the child model.

\begin{table}[htbp]
  \centering
  \caption{Cascade evaluation on 146{,}735 validation samples from fourteen datasets (routing threshold 20 years).}
  \label{tab:cascade-internal}
  \begin{tabular}{lrccc}
    \toprule
    Age group & n & MAE & CS@5 & Routed to child model \\
    \midrule
    0--3 (infants)   & 3{,}162  & 1.925 & 94.1\% & 98.5\% \\
    4--7             & 5{,}504  & 2.296 & 94.6\% & 97.2\% \\
    8--12            & 3{,}389  & 3.211 & 83.9\% & 92.4\% \\
    13--16           & 5{,}315  & 5.430 & 50.4\% & 66.2\% \\
    17--18           & 3{,}501  & 5.839 & 49.3\% & 40.7\% \\
    19--30           & 45{,}753 & 3.956 & 71.9\% & 7.6\% \\
    31--50           & 57{,}134 & 4.008 & 71.0\% & 0.3\% \\
    51--70           & 19{,}313 & 5.295 & 59.8\% & 0.1\% \\
    71+              & 3{,}664  & 9.792 & 27.0\% & 0.1\% \\
    \bottomrule
  \end{tabular}
\end{table}

In the bands the specialist was trained for, accuracy meets the design target with margin: MAE 1.925 for infants, 2.296 for ages 4--7 and 3.211 for ages 8--12, with CS@5 between 84 and 95 per cent.
On its own children-only validation the specialist reaches an MAE of 2.291 for ages 0--18 and 1.944 for ages 0--12, against a baseline estimate of 7--10 years for the unadapted model in that range.
Aggregated over all children (0--18) the cascade improves on the general-only configuration by 0.180 years MAE (3.781 versus 3.960); aggregated over everyone it costs 0.099 years (4.273 versus 4.174), because 7.6 per cent of young adults are misrouted and adolescents suffer under a 0--12 specialist.
The 13--18 band remains the hardest for every configuration we measured, with MAE above 5.4 years; no age-estimation decision at the legal working-age boundary should currently rest on the model alone.

\subsection{Quantisation}
\label{sec:results-quant}

\begin{table}[htbp]
  \centering
  \caption{Age cascade, FP32 PyTorch versus FP8 TensorRT on APPA-Real (batch size 1, NVIDIA GB10).}
  \label{tab:quant-cascade}
  \begin{tabular}{lcc}
    \toprule
    Metric & FP32 & FP8 TRT \\
    \midrule
    MAE (years)          & \textbf{4.755} & 4.757 \\
    Median AE (years)    & 3.725 & \textbf{3.719} \\
    CS@5                 & \textbf{62.9\%} & 62.6\% \\
    CS@10                & \textbf{90.5\%} & 90.4\% \\
    Latency, mean (ms)   & 22.1 & \textbf{12.5} \\
    Latency, p50 (ms)    & 19.5 & \textbf{10.8} \\
    Latency, p99 (ms)    & 39.0 & \textbf{22.7} \\
    \bottomrule
  \end{tabular}
\end{table}

FP8 compilation of the age cascade is essentially free in quality terms (Table~\ref{tab:quant-cascade}): MAE changes by $+0.002$ years (0.04 per cent), per-bucket deltas stay below 0.006 years in either direction, so no age group pays disproportionately, and mean latency drops from 22.1 to 12.5~ms per image, a 1.77$\times$ speedup with corresponding tail-latency gains.

\begin{table}[htbp]
  \centering
  \caption{Full-pipeline ablation of detector precision on the 19-video corpus (482{,}486 frames); all other components fixed.}
  \label{tab:quant-pipeline}
  \begin{tabular}{lccc}
    \toprule
    Metric & FP16 engine & FP8 engine & $\Delta$ \\
    \midrule
    Unique child candidates & \textbf{615} & 602 & $-2.1\%$ \\
    Persons tracked & 40{,}363 & 40{,}047 & $-0.8\%$ \\
    Total processing time (min) & 959.6 & \textbf{810.5} & $-15.5\%$ \\
    Mean throughput (FPS) & 8.9 & \textbf{10.5} & $+18.0\%$ \\
    Mean real-time ratio & 1.78$\times$ & \textbf{2.11$\times$} & $+18.5\%$ \\
    Mean fusion score & \textbf{0.843} & 0.832 & $-1.3\%$ \\
    Gender-reliable share & 74.1\% & \textbf{76.7\%} & $+2.6$ pp \\
    Engine size (MB) & 197 & \textbf{104} & $-47.2\%$ \\
    \bottomrule
  \end{tabular}
\end{table}

At pipeline level (Table~\ref{tab:quant-pipeline}) the FP8 detector engine trades 2.1 per cent of detected child candidates for an 18 per cent throughput gain and half the engine memory.
The loss is not uniform: crowded, low-resolution documentary scenes account for most of it (the most crowded investigation film loses 14 candidates), while several feature films actually gain candidates under FP8.
Age-bucket composition shifts only marginally (mean fused age 11.5 to 11.7 years), and gender reliability improves slightly.
We adopt FP8 as the default study configuration and retain FP16 where recall is critical; Section~\ref{sec:discussion} discusses this operating-point choice.

\subsection{End-to-End Study on the Film Corpus}
\label{sec:results-pipeline}

Table~\ref{tab:pipeline-versions} summarises the study on the 19-video corpus for the previous prototype generation v5.0 (YOLOv8x detector, MiVOLO~v1 at $224\times224$) and the system presented here as v6.1 (YOLO26x CerberusDet, MiVOLO~v2 cascade at $384\times384$).
Both process identical inputs on the same hardware.
Because detector and age model changed together, per-component attribution on this corpus is not possible; the component benchmarks above serve that purpose.

\begin{table}[htbp]
  \centering
  \caption{End-to-end comparison on 19 videos (26.8 h, 482{,}486 frames). The corpus carries no ground-truth annotations; figures describe system yield, not measured recall.}
  \label{tab:pipeline-versions}
  \begin{tabular}{lcc}
    \toprule
    Metric & v5.0 & v6.1 \\
    \midrule
    Unique child candidates & 285 & \textbf{634} ($+122\%$) \\
    Persons tracked & 36{,}773 & 40{,}361 ($+10\%$) \\
    Child candidates under age 10 & 23 & \textbf{213} (9.3$\times$) \\
    Minimum fused age (years) & 6.3 & 0.8 \\
    Mean fusion score & 0.774 & \textbf{0.852} \\
    High-confidence share ($\geq 0.8$) & 49\% & \textbf{69\%} \\
    Share with score $\geq 0.9$ & 24\% & \textbf{52\%} \\
    Records with undetermined gender & 0\% & 25.6\% \\
    Aggregate processing time & \textbf{13.0 h} & 15.9 h \\
    Mean throughput (FPS) & \textbf{11.6} & 8.9 \\
    \bottomrule
  \end{tabular}
\end{table}

The new pipeline surfaces 634 unique child candidates against 285, an increase of 122 per cent, and the confidence structure improves rather than dilutes: the mean fusion score rises from 0.774 to 0.852, and the share of high-confidence records grows from 49 to 69 per cent.
The gain is concentrated exactly where a monitoring system should find it.
The five largest per-video increases all occur in child-labour documentaries, including 6 to 66 candidates on an investigative film about enslaved children and 3 to 47 on a cocoa-sector documentary, while the single regression is a feature film (47 to 43).
Detection of young children changes qualitatively: v5.0 never fused an age below 6.3 years and produced 23 under-10 candidates in 26.8 hours, whereas v6.1 reaches infants (minimum 0.8 years) and produces 213, a 9.3-fold increase, consistent with the specialist's accuracy in the 0--12 band (Section~\ref{sec:results-cascade}).

Two costs accompany the gains.
Throughput falls 23 per cent (8.9 versus 11.6 FPS mean), driven by the 2.9$\times$ larger age-model input and by face re-identification moving to CPU in this build; at 1.7$\times$ aggregate real-time speed (2.1$\times$ with the FP8 detector engine) the system still processes continuous single-camera feeds with headroom.
Gender determination regressed: 25.6 per cent of v6.1 child records carry no gender (v5.0: none), and low-gender-confidence flags doubled.
Track-level quality flags shifted accordingly: extreme per-track age ranges became rarer ($-24$ per cent), while frame-level age jitter and short-visibility flags increased, the latter mostly because the stronger detector now picks up brief appearances that v5.0 missed entirely.
We analyse the gender regression, its likely configuration-level cause and its remediation in Section~\ref{sec:discussion}.

\subsection{Field Pilot in Zimbabwe}
\label{sec:results-field}

The pilot of Section~\ref{sec:eval-method} is, to our knowledge, the first published account of a vision pipeline of this kind operating unattended at a real site with children present, against an external attendance register.
We report it in four parts: deployment scale and stability, bring-up findings, yield against the register, and the measured limits.

\subsubsection{Scale and Stability}

\begin{table}[htbp]
  \centering
  \caption{Field pilot volume over seventeen days (six cameras, one Jetson AGX Thor).}
  \label{tab:field-volume}
  \begin{tabular}{lr}
    \toprule
    Metric & Value \\
    \midrule
    Ten-minute segments processed & 13{,}721 \\
    Frames processed & 38{,}748{,}542 \\
    Person tracks accumulated & 506{,}672 \\
    Segment-level child detections & 3{,}620 \\
    Daily evidence packages produced & 938 \\
    Evidence held for review by the accuracy gate & 48 \\
    Reports filed into the partner case-management system & 1{,}275 \\
    Median per-camera throughput & 5.0 FPS (target 5.0) \\
    Logged errors in the retained log set (31 July--13 August) & 1 \\
    \bottomrule
  \end{tabular}
\end{table}

Table~\ref{tab:field-volume} summarises the volume.
The system held its five-frames-per-second target on all six cameras as a median, processed 38.7 million frames without operator intervention, and filed every queued report successfully (none failed, none stuck in retry).
The single logged error over the whole retained log window (31 July to 13 August) was a camera reader stall that the supervision layer resolved by design: the stall timeout fired, the segment closed cleanly, and the worker restarted.
Crowded scenes on the best-placed camera drop to about 2.1 frames per second, and measurement showed this to be harmless: grouping 3,350 segments by processing time, slow segments yielded three times more children per tracked person than fast ones, because the same crowding that slows the pipeline is where the children are.
Attempting to speed up the per-frame path for recall would have optimised the wrong variable.

\subsubsection{Bring-Up: Software Tuning and Camera Settings}

Detection yield at switch-on was poor and the gap was closed in software within two days: retuning of thresholds and fusion hygiene raised the child-detection rate from 0.005 to 0.181 children per segment, a 36-fold improvement measured across the tuning boundary, confirming an early estimate that the untuned system missed about 98 per cent of visible children.
The single camera-side change that survived measurement was digital wide-dynamic-range (WDR), evaluated with an interleaved A/B design that toggled the setting every twenty minutes: faces per tracked person rose from 8.9 to 11.4--12.2 per cent ($z = 3.4$--$4.6$), median face luminance rose from 32 to 58, and child faces in diagnostic samples rose 8.6-fold.
The interleaving mattered more than the result: a preceding sequential A/B had shown sharpness collapsing by 62 per cent, which interleaving revealed to be an evening-light artefact (the true change was $-6$ per cent).
One caveat is carried forward honestly: brightening shifts the whole age distribution downwards, and after stratifying by luminance and face size a residual shift of $-0.8$ to $-1.8$ years remains, so part of the WDR child-face gain is genuine discovery and part is estimator bias, and the two cannot be separated without labelled field data.

\subsubsection{Yield Against the Attendance Register}

\begin{table}[htbp]
  \centering
  \caption{Daily unique child identities filed versus the attendance register (fourteen register-covered windows). Identity consolidation (Section~\ref{sec:field-methods}) deployed on 4 August.}
  \label{tab:field-roster}
  \begin{tabular}{lcccc}
    \toprule
    Window & Encounters & Identities filed & Children on register & Over-report \\
    \midrule
    31 July & 77 & 44 & 18 & $\times$2.4 \\
    1 August & 359 & 164 & 18 & $\times$9.1 \\
    2 August & 296 & 141 & 19 & $\times$7.4 \\
    3 August & 173 & 104 & 18 & $\times$5.8 \\
    \midrule
    4 August & 202 & 64 & 18 & $\times$3.6 \\
    5 August & 225 & 63 & 18 & $\times$3.5 \\
    6 August & 219 & 42 & 23 & $\times$1.8 \\
    7 August & 234 & 61 & 17 & $\times$3.6 \\
    8 August & 272 & 47 & 17 & $\times$2.8 \\
    9 August & 262 & 66 & 17 & $\times$3.9 \\
    10 August & 278 & 52 & 20 & $\times$2.6 \\
    11 August & 253 & 45 & 17 & $\times$2.6 \\
    12 August & 251 & 45 & 17 & $\times$2.6 \\
    13 August & 298 & 62 & 16 & $\times$3.9 \\
    \bottomrule
  \end{tabular}
\end{table}

Table~\ref{tab:field-roster} shows the pilot's central quality result, covering the fourteen daily windows for which the partner's register is available.
Before consolidation, per-encounter matching over-reported the register by up to a factor of 9.1, one child becoming many filed identities.
The union-find consolidation with the simultaneity veto cut this to between 1.8 and 3.9, compressing identities by a factor of 2.3 to 3.2 per day while the time guard blocked 11 to 56 candidate merges per window.
The register itself stays nearly flat at 16 to 23 children while the over-report factor moves between 1.8 and 3.9, so the day-to-day spread is a property of the system rather than of site activity.
The safety property held throughout: across 240 proven-different encounter pairs and the full threshold grid, the deployed consolidation produced zero false merges.
The counterfactual justifies the veto's existence.
Without it, a slightly more aggressive threshold pair hit the register count exactly, 22 identities on a 22-person day, and was wrong: 69 of its 91 checkable merges joined provably different people.
Matching the target count was over-merging in disguise, and only the impossible-pairs ground truth exposed it.
The same-segment guard closed the complementary failure: before it, six merges had hidden one person inside another's identity, mostly children (two aged 8.0 and 7.9 with fusion scores above 0.99), about 1.4 hidden children per day; after deployment, no defective pairs occurred, at a measured cost of one wrongly split legitimate pair every four days.
The accuracy gate held back 48 evidence items over the pilot, including an aggregated record whose recomputed age of 29.3 years would otherwise have been filed as a child.

Residual over-reporting is carried largely by \emph{singletons}, identities seen in exactly one encounter (47 to 83 per cent of filed identities per day), which current embeddings cannot merge: 64 per cent of them share no body similarity above threshold with anyone, and their median best face similarity is 0.434.
An early explanation went further and did not survive more data, a correction we report deliberately.
Over the first four register-covered windows, singleton share and over-report correlated at $r = +0.90$; over the six windows added by the third register the correlation fell to $+0.20$ ($+0.50$ across all ten), with the highest-singleton day showing one of the lowest over-report factors.
Four points had sufficed to convince us; they were not enough, and the pilot's own rule, validate every candidate conclusion on a second day, applied to thresholds but had not been applied to this explanation.
Singleton share is a contributor, not the mechanism, and the driver of the day-to-day spread remains unidentified.
On the re-identification corpus the deployed lightweight face pack separates same-person from different-person pairs with a gap of 0.298 (61.2 per cent correct merges at a five per cent false rate); a heavier embedder raises the gap to 0.404 and correct merges to 91.2 per cent but finds faces on only 19 per cent of crops against 51 per cent, which is why it was measured and not deployed.
The embeddings, not the thresholds, are the current limit on merging singletons.

\subsubsection{Measured Limits}

Conditional recall, measured over 595 child events in the A/B window, was 48 per cent: of 128 segments in which a child was visible with a detectable face, 61 were confirmed by the pipeline, ranging from 29 to 65 per cent per camera.
This is recall conditional on a face having been found at all; children whose faces were never detected are outside the denominator, so absolute recall is lower.
The dominant limit is camera geometry, not models: the one well-placed camera yields a face for 35 to 39 per cent of tracked people, the other five for 0.0 to 1.9 per cent, and the statistically significant WDR gain in child faces is confined to that single camera ($p = 5\times10^{-14}$ there, $p = 0.41$ elsewhere).
No software change in the pilot moved the other five cameras' yield; remounting them is worth more than any model improvement.

Two further limits define the honest boundary of the method.
The register lists four children aged six or under; the system filed one to two records for them per day.
The detector is not the obstacle, a geometric probe for carried infants (a small box contained within a larger person box at infant proportions) fired 56 times, but neither age path yields information for this band: the body-only path returns its 22.4-year dataset prior in 89 per cent of cases, and across the pilot the face path produced no estimate below seven years for these children, because a carried infant's face is turned away or pressed against the carrier.
There is nothing for thresholds to fix; the measured route is to send the carry geometry itself to human review, a design decision deliberately deferred beyond the pilot.
Finally, the age estimates that drive filing are not humanly verifiable on this footage: child faces from the best camera are 22-to-38-pixel backlit profiles, and contact sheets of neighbouring age bands are visually indistinguishable, so a reviewer cannot confirm or refute an estimate from the evidence crop alone.
On 58 hand-labelled evidence items the deployed recall-first operating point measures precision 0.64 at recall 0.83; a stricter alternative reaches precision 0.90 at recall 0.60 at the identical F1 of 0.72, making the choice between them a policy decision about whose error costs more, and Section~\ref{sec:discussion} argues it must be made by the child-protection partner, not by the system's developers.

\section{Discussion}
\label{sec:discussion}

\subsection{Implications for Monitoring Practice}

The results support a specific, bounded claim: a purpose-built vision pipeline can supply CLMRS programmes with a continuous stream of reviewable child-presence candidates at workplaces, at a quality level that interview-based monitoring cannot reach between visits, and it can do so unattended, on site, for weeks.
The component evidence matters as much as the headline yield.
Detector recall improved by 24 percentage points for persons and 11 for faces, and every recall point removes cases that no downstream component could otherwise recover.
The age benchmark shows that this capability cannot be assembled from default open-source parts: the two stacks a practitioner would most likely deploy mis-age children by 18 to 23 years on average, predicting adults where children stand.
A protection instrument built on such components would not merely fail, it would certify the absence of the children it was meant to find.

The field pilot adds a lesson the benchmarks could not: most of the deployed system's quality was won or lost \emph{above} the models.
The models were frozen throughout the pilot, yet detection yield improved 36-fold through threshold and fusion tuning, over-reporting fell from 9.1 times the register to as low as 1.8 through identity consolidation, and the decisive safety property, zero false merges across 240 proven-different pairs, came from a temporal-logic constraint rather than from any embedding.
The corresponding negative result generalises: a consolidation configuration that exactly matched the register count was over-merging catastrophically, which means yield metrics against an external roster can validate a system only together with a false-merge ground truth, never alone.
We commend the impossible-pairs construction, different people proven by simultaneous visibility, as a labelling-free ground truth available to any multi-camera deployment.

Equally practical are the training-time negative results.
Fine-tuning a strong pretrained age model at conventional rates destroyed it; a validation set that did not match the specialisation target masked genuine progress for a full training stage; aggressive oversampling traded adult accuracy for child accuracy at a ruinous rate.
The stable solution was architectural, a specialist behind a routing threshold, not a data-balancing trick.
The pilot then revised the routing threshold itself (20 to 32 years), because the quantised general model's collapse to its dataset prior on small, backlit faces starved the specialist of exactly the inputs it exists for, a coupling between quantisation, image quality and cascade design that no laboratory benchmark exposed.

The system's place in a CLMRS is as a triage layer.
Its output is a queue of person records ordered by confidence tier, each carrying best-shot evidence, an age estimate with dispersion, and quality flags; field staff verify, dismiss or act.
This division of labour matches how the sector's own effectiveness review describes the identification bottleneck: monitoring visits are episodic and identification rates vary widely across projects \cite{initiativeEffectivenessReviewChild2021}, while survey-based measurement under-reports by design \cite{variousMeasuringChildLabor2025}.
Area-level tools such as GIS risk mapping and satellite monitoring \cite{faoDIGICHILDExploringGeoreferenced2025,labSentinelKilnDBScalableBrick2025} choose where attention should go; our system operates after that choice, at the site itself.
The layers are complementary, not competing.

\subsection{Limitations}
\label{sec:limitations}

\paragraph{What the pilot does and does not establish.}
The film corpus carries no ground truth, so its 122 per cent yield increase is not a recall measurement.
The pilot has ground truth for presence (the register) and for merge correctness (impossible pairs), but not for age: the children on site are known, yet which filed record corresponds to which child is verified only through hand labels on a small sample.
Conditional recall of 48 per cent counts only children whose face was detectable at all, so absolute recall is lower and unmeasured.
Over-reporting of 1.8 to 3.9 means each real child still generates about two to four records per day for reviewers, and the driver of the day-to-day spread in that factor is not yet identified: the singleton explanation that fitted the first four register-covered windows collapsed on the next six (Section~\ref{sec:results-field}).
And a single 17-day pilot at one site, with one camera well placed and five badly, characterises this deployment, not the method's distribution across sites.

\paragraph{Age is unverifiable on field evidence.}
The pilot's most uncomfortable finding is not a model error but an epistemic one: on 22-to-38-pixel backlit profile faces, neighbouring age bands are visually indistinguishable to humans, so the evidence attached to a filed record cannot confirm the age estimate that triggered it.
Filing decisions therefore rest on estimates that neither the reviewer nor the operator can verify from the footage.
A minimum face-size and frontality requirement before filing was identified during the pilot and deliberately left unimplemented pending the partner's policy decision, since it trades recall for verifiability.
Until such a gate or a labelled field set exists, every filed record must be treated as a lead for human investigation, never as evidence of age.

\paragraph{The youngest children are beyond the method.}
The 0--6 band, arguably the highest-priority group, is effectively invisible to face-based age estimation in the field: carried infants are detected as geometry but yield no usable age signal from either the face or the body path.
This is not a tuning gap, there is no information to recover, and we state it as a boundary of the approach.
The measured way forward routes carry geometry directly to human review.

\paragraph{The adolescent band.}
Every configuration we measured degrades at ages 13--19 (MAE above 5.4 years internally; 7.4 for the reference cascade on FairFace-Africa).
The laboratory routing threshold of 20 makes the band worse by sending adolescents to a 0--12 specialist, and the field threshold of 32 routes even more of it there; the cascade improves children 0--18 in aggregate by 0.180 years MAE but costs 0.099 years overall.
Age estimates near legal working-age boundaries therefore carry the least evidential weight precisely where legal consequence is highest, and threshold-level error rates rather than MAE are the correct lens for that band \cite{gaulUnderageDetectionMultiTask2025,variousUnderageDetectionMultiTask2025}.

\paragraph{Gender regression.}
A quarter of v6.1 child records on the film corpus carry no gender determination, against none in v5.0, and gender accuracy (96.2 per cent on specialist validation, 98.7 per cent for the general model) sits below the 99 per cent design target.
The evidence points to a configuration-level cause, a missing abstention-threshold default for the new checkpoint format, and threshold tuning on held-out data is the planned fix.

\paragraph{Cascade versus single model.}
On adult-inclusive public benchmarks the unmodified MiVOLO~v2 edges out the cascade, including on child subsets dominated by teenagers, while the cascade wins decisively in the 0--12 bands (Table~\ref{tab:age-buckets}).
We consider the youngest measurable bands the correct optimisation target for this application, but the trade-off is genuine, and deployments prioritising adolescent screening should weigh the single-model configuration.

\paragraph{Fairness coverage.}
FairFace-Africa results (MAE 5.26 overall, 4.64 on children) show the system functioning on the primary deployment demographic, and the pilot ran at an African site, but a per-skin-type (Fitzpatrick) audit has not yet been run, and the age-estimation literature documents demographic bias as a persistent risk \cite{karkkainenFairFaceFaceAttribute2021,variousEthnicRepresentationMatters2024,cliffordTwoSourcesBias2018}.
The audit is scheduled before any scale-up.
A related field observation cuts both ways: image brightening (WDR) shifted age estimates downwards by 0.8 to 1.8 residual years, which means capture conditions themselves are a bias source that fairness audits on curated crops will not see.

\paragraph{Engineering constraints.}
TensorRT engines are compiled per GPU architecture and per batch size; the field detector runs FP16 rather than FP8 because output-quantised FP8 costs 3 to 5 per cent accuracy on this model; the FP16 TorchScript path is broken for VOLO-family models; and FP4 tooling silently degrades to FP16 kernels on current releases (Section~\ref{sec:quantisation}).
The best detector checkpoint was lost to a backup that continued training over it, and the deployed engine derives from an earlier epoch within 0.5 per cent of it, a mundane failure worth a sentence because provenance discipline for models is as consequential as for data.

\subsection{Ethics, Law and Deployment Safeguards}
\label{sec:ethics}

A system that watches workplaces for children processes the biometric data of minors, the most protected data category in most jurisdictions, and its governance is as important as its accuracy.
Four design commitments follow from the applicable frameworks: the GDPR and the EDPB guidance on video devices and on age assurance \cite{boardEDPBGuidelines32019,boardEDPBStatementAge2025}, the risk-based obligations of the EU AI Act for biometric systems \cite{parliamentEUAIAct2024}, and UNICEF's policy guidance on AI affecting children \cite{unicefPolicyGuidanceAI2024}.

First, processing stays at the edge.
Footage is analysed on site; only fused person records, embeddings and best-shot crops for flagged candidates leave the device, under retention limits, which follows the data-minimisation principle.
The pilot operated this way: the device held its diagnostics locally and the case-management system received only per-identity evidence packages.
Second, the system estimates age; it does not identify.
Re-identification embeddings link appearances pseudonymously and are not matched against any external identity database; the attendance register used for evaluation contains counts and ages, and the comparison happens outside the imaging pipeline.
Third, no automated consequence attaches to any output.
Every record enters a human review queue inside an existing case-management workflow, aligning with the CLMRS core criteria, which place identification, verification and remediation with trained personnel \cite{initiativeCLMRSCoreCriteria2025}.
The pilot exposed how easily this principle erodes in engineering practice: records were filed into a live system under a category label while their age estimates were unverifiable from the attached evidence (Section~\ref{sec:results-field}), which is acceptable in a supervised pilot with a cooperating partner and would not be acceptable at scale.
We record it as a requirement, not an achievement: filing gates must enforce evidential verifiability, not only confidence scores.
Fourth, deployment is restricted to partnerships with child-protection actors under purpose limitation, with audit trails and third-party verification, because the dual-use risk of child-detection technology, its potential repurposing for general surveillance of children, is best controlled institutionally rather than technically.
For the same reason this paper withholds the identities of the pilot site and the child-care organisation.

The error asymmetry carries the final ethical instruction.
A false positive costs verification labour; a false negative is a child left unprotected.
The pilot made this trade-off measurable, precision 0.64 at recall 0.83 versus precision 0.90 at recall 0.60 at identical F1, and the equality of F1 is the point: aggregate scores do not decide such questions.
The operating point belongs to the child-protection partner, set explicitly, with the review workload and the miss rate both on the table.

\subsection{Future Work}
\label{sec:future}

The pilot reordered our priorities.
Camera geometry outranks model work: five of six cameras yielded faces for under two per cent of tracked people, and no software moved that number, so mounting height and tilt are the first lever of any next deployment.
A labelled field validation set is the second, because it is the only way to separate real discovery from estimator bias, to validate filing gates, and to convert yield into recall and precision.
On the model side, the measured priorities are a mixed re-identification pack (the strong embedder behind the permissive face detector, prototyped during the pilot), an adolescent-specialist cascade stage or ordinal heads \cite{caoCORALRankConsistent2020} with routing-threshold optimisation, the per-Fitzpatrick fairness audit, and gender-abstention tuning.
On the workflow side: a verifiability gate before filing (minimum face size and frontality), and routing carried-infant geometry to human review, which is the only measured path into the 0--6 band.
Body-based age cues for face-occluded cases remain the relevant research direction for sites where faces are structurally unavailable \cite{variousGaitbasedAgeEstimation2025,variousCelebFBIFullBody2024}.

\section{Conclusion}
\label{sec:conclusion}

No published system had combined real-time individual child detection, child-grade age estimation, identity persistence and edge deployment into one instrument for child-labour monitoring; area-level risk tools and binary adult--child classifiers left that space open.
This paper presented such a system, its mathematical formulation, and measurements at five levels.
A multi-task YOLO26x-based detector raised person mAP@0.5 from 0.390 to 0.683 over the previous-generation baseline on a 13,537-image benchmark.
A cascaded age estimator, pairing the unmodified MiVOLO~v2 with a child specialist trained through a protocol whose failures we documented alongside its final form, reached an MAE of 1.944 years for ages 0--12 on children-only validation, where widely used open-source alternatives err by 18 years and more.
FP8 TensorRT compilation made the cascade 1.77 times faster at a cost of 0.002 years MAE, keeping the pipeline above real-time speed on embedded Blackwell-class devices without cloud connectivity.
On 26.8 hours of demanding proxy footage, the pipeline surfaced 122 per cent more unique child candidates than its predecessor, with 9.3 times more children under ten and a markedly higher confidence profile.

The field pilot carried the system from benchmarks into reality: seventeen days unattended at a working site in Zimbabwe, 38.7 million frames, one logged error, every report delivered.
Its results reshaped our understanding of where quality lives.
With models frozen, software tuning improved yield 36-fold; identity consolidation under a simultaneity veto cut over-reporting from 9.1 times the attendance register to 1.8--3.9 times at zero proven-false merges; a consolidation variant that exactly matched the register count was exposed as catastrophic over-merging by a labelling-free ground truth built from impossible pairs; and an explanation of the residual over-report spread that fitted four days collapsed when six more arrived, a correction we publish alongside the result it corrects.
The pilot also drew the method's boundaries by measurement: camera geometry, not model capacity, capped face yield on five of six cameras; age estimates on 22-to-38-pixel backlit faces cannot be verified by human reviewers; and carried infants, the 0--6 band, produce no usable age signal at all.

The negative results are part of the contribution, from catastrophic forgetting and validation mismatch in training, through quantisation toolchains that fail loudly (FP16 TorchScript) or silently (FP4), to the field lesson that sequential A/B tests of camera settings produce confidently wrong answers.
The boundaries are equally clear.
Accuracy in the adolescent band does not support decisions at legal working-age thresholds, filed records are leads for human investigation rather than evidence of age, and the system is designed to propose, never to decide: verification and remediation remain with the people who run CLMRS programmes.
The next steps are set by the pilot's measurements, camera remounting, a labelled field validation set, a verifiability gate before filing, and human review of carried-infant geometry, alongside the adolescent-stage and fairness work on the model side.
If those steps hold, continuous, privacy-conscious machine observation can become a standard evidence layer in the monitoring systems the sector already trusts, and children who today become visible only when an interviewer calls will be seen in time to be helped.

\newpage
\printbibliography[title=References]

\newpage
\appendix
\section{Per-Video End-to-End Results}
\label{app:per-video}

Table~\ref{tab:per-video} lists per-video results for the end-to-end study of Section~\ref{sec:results-pipeline}.
The corpus mixes investigative documentaries on child labour with feature films and control material; titles are given as identifiers of the source recordings.

\begin{table}[htbp]
  \centering
  \caption{Unique child candidates per video, pipeline v5.0 versus v6.1 (identical inputs and hardware).}
  \label{tab:per-video}
  \begin{tabular}{lcccc}
    \toprule
    Video & Duration & v5.0 & v6.1 & $\Delta$ \\
    \midrule
    Slavery: A Global Investigation        & 1:17:48 & 6  & 66 & $+60$ \\
    The Gods Must Be Crazy                  & 1:48:50 & 3  & 48 & $+45$ \\
    Shady Chocolate                         & 0:44:59 & 3  & 47 & $+44$ \\
    The Dark Truth of Chocolate             & 0:22:28 & 6  & 34 & $+28$ \\
    FLA child-labour mitigation project     & 0:26:22 & 9  & 31 & $+22$ \\
    The Chocolate War                       & 0:57:46 & 6  & 25 & $+19$ \\
    The Dark Side of Chocolate              & 0:46:23 & 3  & 22 & $+19$ \\
    Inside the Life of an 8-Year-Old Labourer & 0:49:48 & 14 & 30 & $+16$ \\
    Child labour department takes action    & 0:44:30 & 4  & 20 & $+16$ \\
    Harry Potter and the Half-Blood Prince  & 2:39:11 & 13 & 29 & $+16$ \\
    The Boy in the Striped Pyjamas          & 1:34:29 & 24 & 38 & $+14$ \\
    Harry Potter and the Sorcerer's Stone   & 2:38:50 & 41 & 54 & $+13$ \\
    Uncovering 8-year-old children working  & 0:21:23 & 11 & 24 & $+13$ \\
    The Gods Must Be Crazy II               & 1:37:30 & 8  & 20 & $+12$ \\
    Jurassic Park                           & 2:06:36 & 19 & 30 & $+11$ \\
    L\'eon                                  & 2:12:54 & 21 & 26 & $+5$  \\
    Charlie and the Chocolate Factory       & 1:55:25 & 25 & 25 & $0$   \\
    Home Alone 2                            & 2:00:01 & 22 & 22 & $0$   \\
    Home Alone                              & 1:42:54 & 47 & 43 & $-4$  \\
    \midrule
    Total                                   & 26.8 h  & 285 & 634 & $+349$ \\
    \bottomrule
  \end{tabular}
\end{table}

\section{Detector Training Configuration}
\label{app:detector-config}

Table~\ref{tab:detector-config} lists the full training configuration of the multi-task detector (Section~\ref{sec:detector}).
Augmentation values were obtained by evolutionary hyperparameter search.

\begin{table}[htbp]
  \centering
  \caption{YOLO26x-pf (CerberusDet) training configuration.}
  \label{tab:detector-config}
  \begin{tabular}{ll}
    \toprule
    Parameter & Value \\
    \midrule
    Hardware & 4$\times$ NVIDIA H200, DDP, SyncBN, AMP \\
    Input size & $640\times640$ \\
    Batch size & 192 (48 per GPU) \\
    Epochs & 166 of 300 (early stop; best epoch 115; patience 50) \\
    Wall time & 71.7 h \\
    Optimiser & SGD, lr$_0$ 0.00309, one-cycle to 0.0956$\times$lr$_0$ \\
    Momentum / weight decay & 0.952 / $3.7\times10^{-4}$ \\
    Warm-up & 2.04 epochs \\
    Loss weights (per task) & box 7.5, cls 0.5, dfl 1.5 \\
    Mosaic / MixUp & 1.0 / 0.285 \\
    Scale / shear / translate / rotate & 0.846 / 0.717 / 0.211 / 0.299 \\
    Flip left--right / up--down & 0.5 / 0.00983 \\
    HSV (h / s / v) & 0.0124 / 0.696 / 0.287 \\
    Blur / median blur / greyscale & $p=0.1$ / $p=0.1$ / $p=0.01$ \\
    \bottomrule
  \end{tabular}
\end{table}

\section{Age-Model Fine-Tuning Phases}
\label{app:age-phases}

Table~\ref{tab:age-phases} summarises the successful fine-tuning phases of both cascade members (Section~\ref{sec:training-protocol}); validation MAE is measured on the validation distribution matching each phase's training scope.
The failed strategies (high learning rate, pseudo-labelling, self-distillation, mismatched validation, tenfold oversampling) are described in the main text.

\begin{table}[htbp]
  \centering
  \caption{Fine-tuning phases of the general model and the child specialist.}
  \label{tab:age-phases}
  \begin{tabular}{llcccc}
    \toprule
    Model & Phase & Epochs & LR & Scope & Val. MAE \\
    \midrule
    General & warm-up (frozen backbone) & 5 & $3\times10^{-5}$ & all ages & 5.039 $\to$ 4.785 \\
    General & clean fine-tune           & 40 & $2\times10^{-6}$ & 8 datasets & $\to$ 3.729 \\
    General & diverse fine-tune         & 30 & $1\times10^{-6}$ & 13 datasets & $\to$ 3.936 \\
    \midrule
    Specialist & warm-up (frozen backbone) & 5 & $5\times10^{-5}$ & ages 0--18 & 3.485 $\to$ 3.050 \\
    Specialist & children fine-tune        & 40 & $5\times10^{-6}$ & ages 0--18 & $\to$ 2.291 \\
    Specialist & infant focus              & 15 & $1\times10^{-6}$ & ages 0--12 & $\to$ 1.944 \\
    \bottomrule
  \end{tabular}
\end{table}

\section{Reproducibility Note}
\label{app:repro}

Detector training used the public CerberusDet framework and public datasets; the model card, training command, architecture definition and hyperparameter files are retained internally and summarised in Appendix~\ref{app:detector-config}.
Age-model fine-tuning used the public MiVOLO codebase with the modifications described in Section~\ref{sec:training-protocol} (children-only validation filtering, distributed weighted sampling).
Benchmark evaluations used pycocotools and the public APPA-Real and FairFace validation splits.
The trained weights and the pipeline implementation are research prototypes, created and used solely to study the feasibility of the approach described here.
Both are intended for public release under an open licence; until the release is complete, researchers can request evaluation access from the author for verification purposes.

\end{document}